\documentclass{article}

\usepackage{microtype}
\usepackage{graphicx}
\usepackage{subfigure}
\usepackage{booktabs} 

\usepackage{amsfonts}
\usepackage{bbm}
\usepackage{physics}
\usepackage{multicol}
\usepackage{makecell}
\usepackage{epstopdf}
\usepackage{tikz}

\usepackage{adjustbox}

\newcommand{\R}{\mathbb{R}}

\usepackage{listings}
\usepackage{float}
\newfloat{lstfloat}{htbp}{lop}
\floatname{lstfloat}{Listing}

\usepackage{hyperref}

\usepackage[accepted]{icml2022}

\icmltitlerunning{Few-Bit Backward: Quantized Gradients of Activation Functions for Memory Footprint Reduction}

\definecolor{codegreen}{rgb}{0,0.6,0}
\definecolor{codegray}{rgb}{0.5,0.5,0.5}
\definecolor{codepurple}{rgb}{0.58,0,0.82}
\definecolor{backcolour}{rgb}{0.95,0.95,0.92}

\lstdefinestyle{mystyle}{
    backgroundcolor=\color{backcolour},
    commentstyle=\color{codegreen},
    keywordstyle=\color{magenta},
    numberstyle=\tiny\color{codegray},
    stringstyle=\color{codepurple},
    basicstyle=\ttfamily\footnotesize,
    breakatwhitespace=false,
    breaklines=true,
    captionpos=b,
    keepspaces=true,
    numbers=left,
    numbersep=5pt,
    showspaces=false,
    showstringspaces=false,
    showtabs=false,
    tabsize=2
}
\hypersetup{
    pdfsubject={Machine Learning, Deep Learning, Artificial Intelligence},
}

\begin{document}

\twocolumn[
    \icmltitle{Few-Bit Backward: Quantized Gradients of Activation Functions for Memory Footprint Reduction}
    \icmlsetsymbol{equal}{*}

    \begin{icmlauthorlist}
        \icmlauthor{Georgii Novikov}{skoltech}
        \icmlauthor{Daniel Bershatsky}{skoltech}
        \icmlauthor{Julia Gusak}{skoltech}
        \icmlauthor{Alex Shonenkov}{sber}
        \icmlauthor{Denis Dimitrov}{sber,msu}
        \icmlauthor{Ivan Oseledets}{skoltech,airi}
    \end{icmlauthorlist}

    \icmlaffiliation{airi}{AIRI, Moscow, Russia}
    \icmlaffiliation{msu}{Lomonosov MSU, Moscow, Russia}
    \icmlaffiliation{sber}{Sber AI, Moscow, Russia}
    \icmlaffiliation{skoltech}{Skoltech, Moscow, Russia}

    \icmlcorrespondingauthor{Georgii Novikov}{georgii.novikov@skoltech.ru}

    \icmlkeywords{Machine Learning, Transformers}

    \vskip 0.3in
] 



\printAffiliationsAndNotice{}  

\begin{abstract}
Memory footprint is one of the main limiting factors for large neural network training. In backpropagation, one needs to store the input to each operation in the computational graph. Every modern neural network model has quite a few pointwise nonlinearities in its architecture, and such operation induces additional memory costs which~--- as we show --- can be significantly reduced by quantization of the gradients.
We propose a systematic approach to compute optimal quantization of the retained gradients of the pointwise nonlinear functions with only a few bits per each element.
We show that such approximation can be achieved by computing optimal piecewise-constant approximation of the derivative of the activation function, which can be done by dynamic programming. The drop-in replacements are implemented for all popular nonlinearities and can be used in any existing pipeline. We confirm the memory reduction and the same convergence on several open benchmarks.
\end{abstract}

\section{Introduction}

    \begin{figure}[t] \label{fig:3bit-gelu}
        \vskip 0.2in
        \centering
        \input{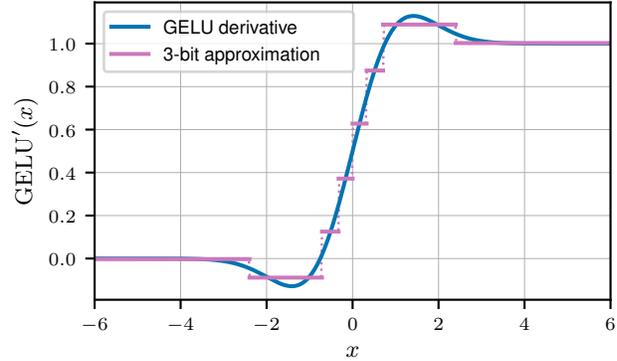}
        \vskip -0.2in
        \caption{Optimized 3-bit piecewise-constant approximation of the derivative of the GELU activation function.}
    \end{figure}

    Modern neural networks models are getting larger and larger. One of the main bottlenecks in the training loop is the required device memory storage \cite{ojika2020addressing,gao2020estimating}. In this paper, we propose a universal approach that helps to reduce the model memory footprint during backpropagation. Note that this approach is complementary to other memory reducing techniques such as checkpointing \cite{Chen16} or offloading \cite{Beaumont21}. Our method can be applied to any neural network without any additional preprocessing.

    The memory consumed by the model during training (except intermediate tensors) can be split into two groups: 1) the model weights (including additional memory for the optimizer state), 2) activations saved for the backward pass, over which the computation is not carried out directly at the moment, but which will be required in the future to compute the gradients.

    Every operation in the computational graph generates a memory footprint. It is typically overlooked, that the application of the pointwise non-linearity (such as GELU or sigmoid) actually results in storing the input for the backward pass. We show that instead of keeping the full input tensor, it is possible to store a low-bit representation, which allows accurate gradients approximation.
.

    In this work, we propose to approximate the derivative of the activation function in a piecewise-constant form. Such an  approximation problem has to be solved once for each activation function, and we propose a simple technique to do that.

    The proposed approximation divides all values into several bins and saves only its corresponding bin indices instead of storing all values. This is a lossy compresion, but the additional noise introduced by it is negligible as we will show on several benchmarks~\ref{sec:experiments}.

    Main contributions of our paper are:
    \begin{itemize}
       \item We propose new approximate backward computation schemes that significantly reduce the memory consumption of neural network training.
       \item We benchmark our approach on several tasks. We show that it provides up to 40\% memory reduction on various tasks while maintaining the accuracy on par with the model trained via the standard approach
   \end{itemize}

\section{Quantized Gradients of Activations}
\paragraph{Gradients of activations using automatic differentiation.}
    \begin{figure}[h] \label{fig:calc-graph}
        \centering
        \includegraphics[width=0.52\textwidth]{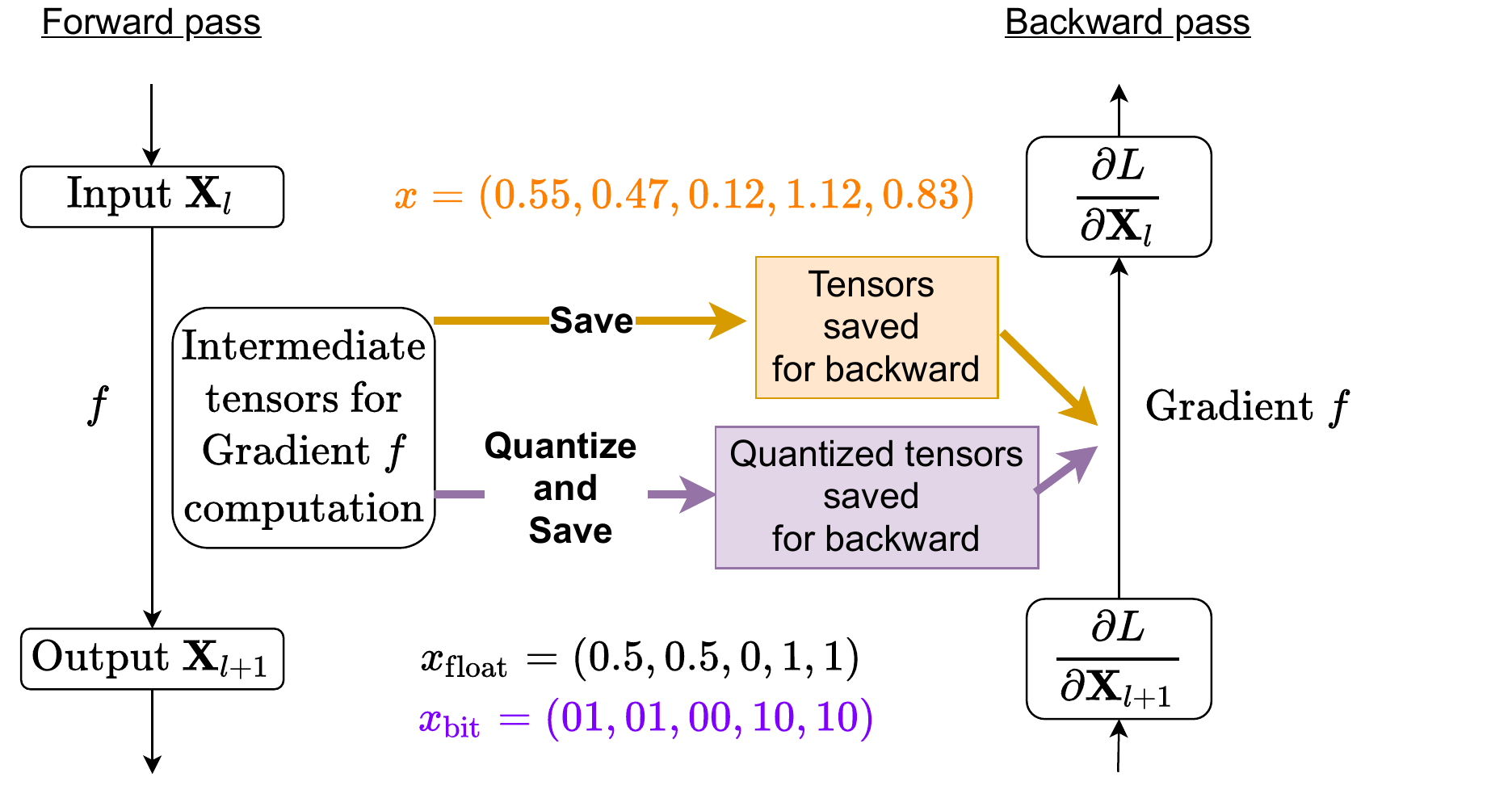}
        \caption{Computation graph of both forward and backward pass. Orange and purple parts of the graph correspond to standard and proposed ways of saving tensors for backward, respectively. Vector $x_{\mathrm{bit}}$ stands for the tensor saved using 2-bit quantization, while $x$ denotes its uncompressed version.}
    \end{figure}

    Modern deep learning frameworks use the \emph{reverse mode automatic differentiation}  to calculate the gradients of the loss over the model parameters. Forward computation can be associated with a directed acyclic graph, depicted in Fig.~\ref{fig:calc-graph}. Each operation $f$ computes the output $\vb{X}_{l + 1}$ given the input $\vb{X}_l$ and has to save some information $\vb{S}_l$ that would be used on the backward pass in order to calculate the derivative $\partial{L} / \partial{\vb{X}_l}$ from $\partial{L} / \partial{\vb{X}_{l + 1}}$ and $\vb{S}_l$. Thus, in a typical training loop, the intermediates $\vb{S}_l$ of all operations in the graph are stored in the memory during the whole forward pass until they are no longer needed after the completion of the corresponding backward operation during backward pass. This generates an additional memory, which can be quite significant and be larger than the total amount of parameters of the model.
    \paragraph{Pointwise activations.}
    In this paper, we focus on a pointwise activation function, which is ubiquitous in modern neural network architectures. Given an input tensor $\vb{X}_l$ we apply a function $f$ to each of the elements of this tensor:

    \begin{equation*}
        f(\vb{X}_l) = [f(\vb{X}_l^{j_1, \dots, j_k})]_{j_1, \dots, j_k}, f : \R \rightarrow \R.
    \end{equation*}

    This operation is very cheap compared to other operations in the deep neural network model and does not attract much attention when analysing computational complexity. However, standard implementation in such a framework as PyTorch induces not a very small memory footprint and the whole input $\vb{X}_l$ is saved for the backward pass.

    The backward pass for such a function consists of element-wise multiplication of the propagated gradient tensor by the derivative of the nonlinearity function at the points of the input tensor: if $\vb{X}_{l+1} = f(\vb{X}_{l})$, then the gradient of the loss $L$ with respect to $\vb{X}_{l}$ is computed as
    \begin{equation}\label{fewbit:chain}
    \frac{\partial L}{\partial \vb{X}_{l}} = \frac{\partial L}{\partial \vb{X}_{l+1}} f'(\vb{X}_{l}),
   \end{equation}
   where $f'(\vb{X}_{l})$ is the tensor with elements, consisting of the derivative of $f$ evaluated in each element of $\vb{X}_{l}$. From \eqref{fewbit:chain}, it follows that for the backward pass we have to store only $f'(\vb{X}_{l})$, and $\vb{X}_{l}$ is not needed.
   \paragraph{ReLU activation function.}
   To illustrate our idea, consider one of the most popular nonlinearities, $f(x) = \mathrm{ReLU}(x) = \max(0, x).$ Its derivative $f'$ takes only two values, $0$ and $1$ and it only require $1$ bit to store. If single precision is used, then the compression is $32$, which is quite noticeable.

   \paragraph{GELU activation function.}
   In modern transformer architectures \cite{Vaswani17} the GELU \cite{Hendrycks16} nonlinearity is typically used. The derivative no longer takes two values. Instead, we propose to approximate $f'$ by a \emph{piecewise-constant} function. For example, if we allow $8$ different values, we will need only $3$ bits per each element (Fig.~\ref{fig:3bit-gelu}).

    {\bf Quantized gradients of activations.} In stochastic optimization, if the gradient for a given batch is computed approximately, the optimization may still converge.
    The GELU derivative (see Fig.~\ref{fig:3bit-gelu}) is quite ``similar'' to a piecewise-constant approximation: for large values of $x$, it is almost exactly equal to $0$ or $1$,
    and for small values of $x$, a rather interesting transition from $0$ to $1$ occurs. Instead of calculating the derivative exactly on the backward pass, we approximate it using a certain piecewise-constant approximation:
    \begin{equation} \label{eq:approx}
        \mathrm{q}(x | \vb{s}, \vb{y}) = \sum_i y_i \mathbbm{1}[x \in [s_i; s_{i + 1}]]
    \end{equation}
    As noted above, if the approximation has $k$ constant intervals, instead of storing the full input tensor $X$, it will be possible to save only $\log k$ bits of information
    (per element of the input tensor), which, accordingly, will reduce the memory consumption by $32 / \log k$ times for single precision.

    Fig.~\ref{fig:3bit-gelu} shows an example of an optimized 3-bit piecewise-constant approximation for the GELU activation function.
    Finding the optimal approximation parameters (boundaries of intervals and values on them) is a challenging task. We propose to find them by minimizing the (weighted) $L_2$ norm of the error.

\section{Optimal Piecewise-constant Approximation}

    \begin{figure*}[h] \label{fig:3-bit-examples}
        \centering
        \includegraphics[width=0.9\textwidth]{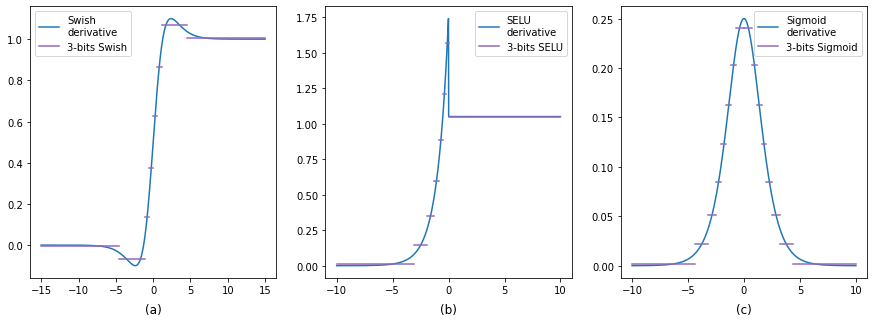}
        \caption{Examples of 3-bit approximations for derivatives of popular nonlinearities: (a) Swish, (b) SELU, and (c) Sigmoid. Please note that the derivative of the sigmoid is an even function, thus we can quantize only positive real axis $[0; +\inf]$, which essentially doubles the memory budget.}
    \end{figure*}

	Consider function $f: \mathbb{R} \rightarrow \mathbb{R}$ and its derivative $f'$. We will measure the quality of a piecewise constant approximation~\eqref{eq:approx} with a weighted $L_2$ norm
	\begin{equation} \label{eq:approx-quality}
		\min_{\vb{y}, \vb{s}} L(\vb{s}, \vb{y}), \, L(\vb{s}, \vb{y}) = \int_\R (f'(x) - q(x | \vb{s}, \vb{y}))^2 w(x) dx,
	\end{equation}
	where $w$ is some weight function reflecting our prior knowledge of the activation function argument distribution. Practical choices of $w$ may be either $\mathbbm{1}[x \in [A; B]]$ (with some reasonable $A$ and $B$, which should be large enough) which makes integral \eqref{eq:approx-quality} tractable, or maybe, e.g., standard normal distribution.

	It is easy to see that the optimal value of $\vb{y}$ for $L(\vb{s}, \vb{y})$ with given $\vb{s}$ is:
	\begin{equation} \label{eq:y_opt}
		y_i(\vb{s}) = \frac{\int_{s_i}^{s_{i + 1}} w(x) f'(x) dx}{\int_{s_i}^{s_{i + 1}}w(x) dx}.
	\end{equation}
	The gradient of $L(\vb{s}, y(\vb{s}))$ w.r.t. the vector $\vb{s}$ then can be derived analytically:
	\begin{equation}
	\begin{split}
		& \frac{\partial L}{\partial s_i} = (2f'(s_i) - y_{i}(\vb{s}) - y_{i-1}(\vb{s})) (y_i(\vb{s}) - y_{i - 1}(\vb{s})) w(s_i).
		\end{split}
	\end{equation}
	Using this formula, $L(\vb{s}, \vb{y})$ can be optimized using any gradient-based method, and optimal piecewise-constant approximations can be found for the different number of bits using standard optimization techniques.

	\paragraph{Dynamic programming.}
	The minimization problem \eqref{eq:approx-quality} has many local minima that are far from optimal. We suggest using dynamic programming to get some good initial approximation that can be finetuned using gradient-based methods (but also can be used as is because it is very accurate on its own).

	We will assume that the weighting function $w$ is chosen such that $w(x) = 0$ for $x \not\in [A; B]$. Consider an auxiliary value
	\begin{equation*}
	    \begin{split}
	        & \mathrm{DP}(t, k) = \min_{\substack{y_{1:k}, \\ s_{1:k + 1}, \\ s_{1} = A, \\ s_{k + 1} = t}} \int_{A}^{t} (f'(x) - q(x | \vb{y}, \vb{s}))^2 w(x) dx, \\
	        & t \in \R, k \in \mathbb{N}.
	    \end{split}
	\end{equation*}
	Essentially, $\mathrm{DP}(t, k)$ is the optimal piecewise constant approximation of size $k$ for the given function $f'$ on the interval $[A; t]$. The recurrent formula for this value is:
	\begin{equation} \label{eq:dpcontinuous}
	\begin{split}
	    &\mathrm{DP}(t, k + 1) = \min_{t'} \mathrm{DP}(t', k) + \\
	    & + \int_{t'}^t (f'(x) - y(t', t))^2 w(x) dx, \\
	    &y(t', t) =\int_{t'}^{t} w(x) f'(x) dx,
	\end{split}
	\end{equation}
    since a piecewise-constant approximation of size $k + 1$ consists of corresponding approximation of size $k$ (first term) plus one constant interval (second term). Here $t'$ chooses the right bound of approximation of size $k$, and $y(t', t)$ stands for the optimal value for the interval $[t'; t]$~\eqref{eq:y_opt}. Then the minimal value of $L(\vb{s}, \vb{y})$ of size $k$ is equal to $\mathrm{DP}(B, k)$.

    To solve the minimization problem~\eqref{eq:dpcontinuous}, we suggest considering the discretization of $t$: $A = t_0 < t_1 < \cdots < t_n = B$ and reducing the calculation of $\mathrm{DP}(t, k)$ to its approximation only in the points of discretization:
	\begin{equation} \label{fewbit:dynprog}
	\begin{split}
	    &\mathrm{DP}(i, k) =  \min_{j} \mathrm{DP}(j, k) + T(j, i),\\
	    &T(j, i)  =  \int_{t_j}^{t_i} (f'(x) - y(j, i))^2 w(x) dx, \\
	    &y(j, i)  =  \frac{\int_{t_j}^{t_i} w(x) f'(x) dx}{\int_{t_j}^{t_i} w(x) dx}.
	    \end{split}
	\end{equation}
	Both $y(j, i)$ and $T(j, i)$ can be calculated in advance using analytical formulas (if possible) or numerically for the corresponding 1-dimensional integrals. After that, the full array of $\mathrm{DP}(i, k)$ can be calculated in $\mathcal{O}(n^2 K)$ time and $\mathcal{O}(n^2)$ space, where $K$ is the required number of constant intervals in the approximation~\eqref{eq:approx}. Please note that this optimization has to be performed only once, so $n$ can be chosen quite large thus the result would be very close to the global minimum.

	Note that the space complexity can be reduced to $\mathcal{O}(n)$ by rewriting~\eqref{fewbit:dynprog} as
	\begin{equation}
	\begin{split}
	    &F^2(i) = \int_{A}^{t_i} f'^2(x) w(x) dx, \\
	    &W(i) = \int_{A}^{t_i} w(x) dx, \\
	    &FW(i) = \int_{A}^{t_i} f'(x) w(x) dx, \\
	    &y(j, i) = (FW(j) - FW(i)) / (W(j) - W(i)), \\
	    &T(j, i)  = F^2(i) - F^2(j) - y(j, i)^2  (W(i) - W(j)).
	    \end{split}
	\end{equation}
	We can see that ultimately only $\mathcal{O}(n)$ one-dimensional integrals have to be stored, and everything else can be easily evaluated in $\mathcal{O}(1)$ time on the spot. The one-dimensional integrals can be calculated numerically in $\mathcal{O}(n)$ time and space complexity as well:
	\begin{equation}
	\begin{split}
	    &F^2(i + 1) = F^2(i) + \int_{t_i}^{t_{i + 1}} f'^2(x) w(x) dx, \\
	     & W(i + 1) = W(i) + \int_{t_i}^{t_{i + 1}} w(x) dx, \\
	     &FW(i + 1) = FW(i) + \int_{t_i}^{t_{i + 1}} f'(x) w(x) dx.
	     \end{split}
	\end{equation}

	\paragraph{Numerical results.} In Fig.~\ref{fig:3-bit-examples}, we provide some 3-bit examples for popular activation functions obtained with described method, and in Table~\ref{table:optimal-approximations}, we provide numerical values of error~\eqref{eq:approx-quality} with uniform weight on interval $[-10; 10]$:
	\begin{equation} \label{eq:uniform-weight}
	    w(x) = \begin{cases} 1, \text{ if } x \in [-10; 10] \\ 0, \text{ otherwise} \end{cases}.
    \end{equation}

    Note that the convergence is quite fast with respect to the number of bits: the error drops by a factor of $2-4$ when an additional bit is added. It would be interesting to study the convergence rates of such approximations.
	\begin{table}[h] \label{table:optimal-approximations}
    \begin{tabular}{|l|l|l|l|l|}
    \hline
            & 1-bit & 2-bits & 3-bits & 4-bits \\ \hline
        ReLU & 0.0 & - & - & - \\ \hline
        GELU & 0.1410 & 0.0406 & 0.0119 & 0.0031 \\ \hline
        Swish & 0.2150 & 0.0479 & 0.0170 & 0.0045 \\ \hline
        Sigmoid & 0.0181 & 0.0038 & 0.0009 & 0.0002 \\ \hline
        Tanh & 0.1584 & 0.0319 & 0.0073 & 0.0017 \\ \hline
        SELU & 0.2554 & 0.1010 & 0.0184 & 0.0039 \\ \hline
        Softplus & 0.2902 & 0.0541 & 0.0121 & 0.0029 \\ \hline
    \end{tabular}
    \caption{Numerical values of error~\eqref{eq:approx-quality} with uniform weight on interval [-10; 10]~\eqref{eq:uniform-weight}.}
    \end{table}

	With precalculated piecewise-constant approximation, drop-in replacement for activation function $f$ is very straightforward. On the forward pass, instead of the full tensor $\vb{X}$, we have to save only indices of intervals to which the elements of $\vb{X}$ belong (which would take $\log k$ bits where $k$ is the number of intervals), and on the backward pass, we need to multiply gradient w.r.t. output not with the actual derivative of $f$, but with values from $\vb{y}$ corresponding to stored indices. Pseudocode is presented in Alg.~\ref{alg:forward-backward}.

	\begin{lstlisting}[
	    language=Python,
	    label={alg:forward-backward},
	    caption={Pseudo code for quantized backward layer. Arrays $\vb{s}$ and $\vb{y}$ are parameters of quantization~\eqref{eq:approx}, $\mathrm{sortedsearch}$ is a binary search method.}
	]
	# Globally stored
	# piecewise-constant
	# approximation parameters
	s = [...]
	y = [...]

	def forward(X):
	    X_pos = sortedsearch(s, X)
	    save_for_backward(X_pos)
	    return f(X)

	def backward(dLdY):
	    X_pos = get_saved_for_backward()
	    return dLdY * y[X_pos]
	\end{lstlisting}

    \begin{figure*}[h]
        \centering
        \includegraphics[width=.9\textwidth]{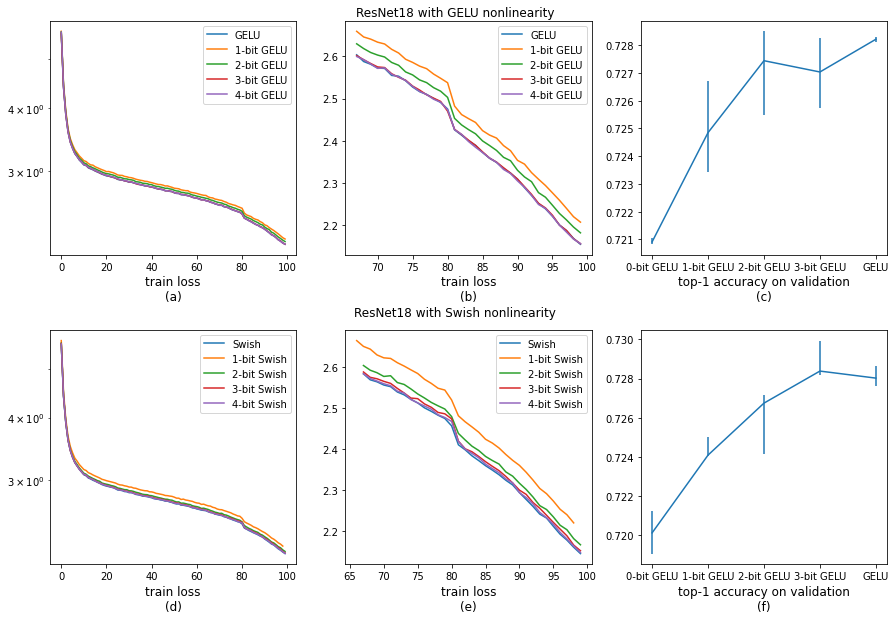}
        \caption{ResNet18 with ReLU replaced with either GELU (a, b, c) or Swish (d, e, f) nonlinearity trained on Imagenet. (a,d): Training loss. (b,e): Training loss during the last third of the training process. (c,f): Final validation top-1 accuracy. All plots are averaged across three runs with different seeds. Error bars mean minimum and maximum values.}
        \label{fig:resnet18-gelu-swish}
    \end{figure*}

    \begin{figure*}[h]
        \centering
        \includegraphics[width=\textwidth]{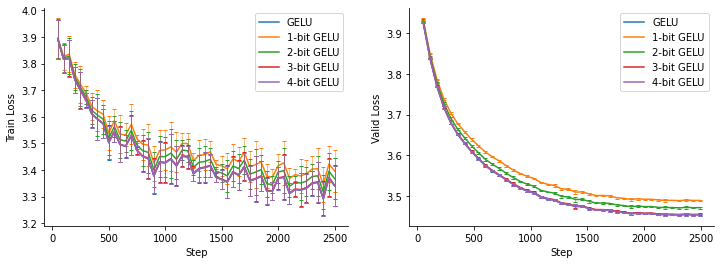}
        \caption{
            Dynamic of loss values in finetuning of ruDALL-E Malevich with few-bit GELU activations. All experiments have following setup: train size 2474, valid size 275, loss image weight 1000, frozen MLP and attention layers, batch size $40$, start lr 4e-7, max lr 1e-5, final lr 2e-8, warmup 0.1, 8bit-Adam~\cite{Dettmers21}, weight decay 0.2, betas (0.9, 0.98), eps 1e-6, gradient checkpointing 24, trained for 6h using 1xA100.
        }
        \label{fig:ru-dalle}
    \end{figure*}

\section{Experiments} \label{sec:experiments}
	The goal of our experiments is to show that quantized gradients give memory savings with no quality degradation. We evaluate our method on language models on several open benchmarks. As Few-bit backward method is a drop-in replace solution, no hyperparameter optimization was performed; all hyperparameters are the same across compared methods and are taken from the corresponding sources.

	In Table~\ref{table:roberta-base-glue} we report results for RoBERTa-base model~\cite{Liu19} on GLUE benchmark~\cite{Wang19} for standard GELU and 1-, 2-, 3- and 4-bits GELU. Both task-relevant metrics and the final value of the loss function are reported. 1- and 2-bits versions have minor performance degradation, while 3- and 4-bits GELU have no visible difference and closely match vanilla GELU performance both in terms of performance metric and loss function while saving $15\%$ and $14\%$ memory respectively~\ref{table:glue-memory}.

    \begin{table*}[t] \label{table:roberta-base-glue}
    \caption{RoBERTa-base on GLUE benchmark with different quantization budgets. Metric: mean accuracy/correlation (task specific). Averaged across five runs.}
    \centering
    \begin{tabular}{llllll}
        & \multicolumn{1}{c}{\textbf{1-bit GELU}} & \multicolumn{1}{c}{\textbf{2-bits GELU}} & \multicolumn{1}{c}{\textbf{3-bits GELU}} & \multicolumn{1}{c}{\textbf{4-bits GELU}} & \multicolumn{1}{c}{\textbf{Vanila GELU}} \\ \hline
        stsb & 0.906 ($\pm$ 0.002) & 0.907 ($\pm$ 0.002) & 0.910 ($\pm$ 0.002) & 0.909 ($\pm$ 0.002) & 0.909 ($\pm$ 0.001) \\ \hline
        mnli-mm & 0.870 ($\pm$ 0.001) & 0.870 ($\pm$ 0.002) & 0.871 ($\pm$ 0.002) & 0.870 ($\pm$ 0.001) & 0.871 ($\pm$ 0.002) \\ \hline
        mrpc & 0.880 ($\pm$ 0.009) & 0.884 ($\pm$ 0.008) & 0.884 ($\pm$ 0.007) & 0.885 ($\pm$ 0.008) & 0.882 ($\pm$ 0.005) \\ \hline
        cola & 0.595 ($\pm$ 0.016) & 0.580 ($\pm$ 0.014) & 0.596 ($\pm$ 0.015) & 0.607 ($\pm$ 0.014) & 0.604 ($\pm$ 0.013) \\ \hline
        mnli & 0.873 ($\pm$ 0.001) & 0.872 ($\pm$ 0.002) & 0.874 ($\pm$ 0.001) & 0.874 ($\pm$ 0.002) & 0.874 ($\pm$ 0.001) \\ \hline
        sst2 & 0.939 ($\pm$ 0.003) & 0.938 ($\pm$ 0.003) & 0.941 ($\pm$ 0.004) & 0.941 ($\pm$ 0.003) & 0.943 ($\pm$ 0.002) \\ \hline
        rte & 0.752 ($\pm$ 0.021) & 0.756 ($\pm$ 0.023) & 0.780 ($\pm$ 0.014) & 0.771 ($\pm$ 0.025) & 0.771 ($\pm$ 0.017) \\ \hline
        qqp & 0.914 ($\pm$ 0.001) & 0.915 ($\pm$ 0.000) & 0.916 ($\pm$ 0.001) & 0.916 ($\pm$ 0.001) & 0.916 ($\pm$ 0.001) \\ \hline
        qnli & 0.925 ($\pm$ 0.002) & 0.925 ($\pm$ 0.002) & 0.926 ($\pm$ 0.002) & 0.927 ($\pm$ 0.002) & 0.927 ($\pm$ 0.002) \\ \hline
    \end{tabular}
    \end{table*}

    \begin{figure*}[h] \label{fig:qqp-losses}
        \centering
        \includegraphics[width=0.9\textwidth]{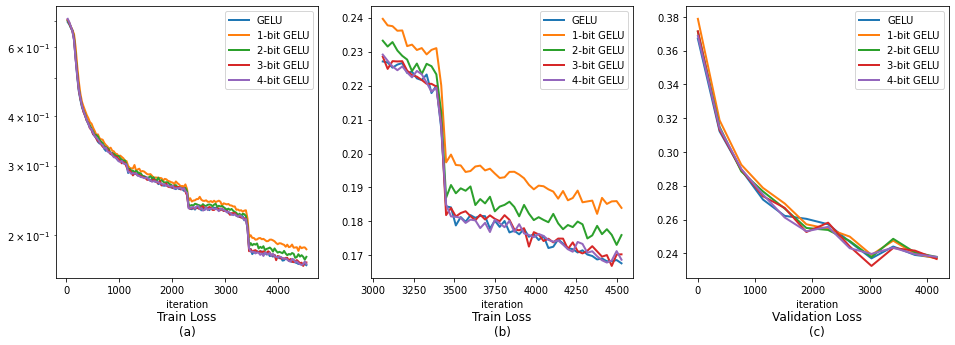}
        \caption{RoBERTa-base on QQP task from GLUE benchmark. (a): Train loss. (b): Train loss during the last third of the training. (c): Validation loss. Averaged across 10 runs.}
    \end{figure*}

    To further examine the influence of backward GELU quantization on stochastic optimization, we consider the behaviour of loss function during training, depicted in Fig.~\ref{fig:qqp-losses}. We can clearly see that our method repeats the dynamic of standard GELU both on train and validation sets. 1- and 2-bit versions performs a little worse, while 3- and 4-bit versions are hardly distinguishable from the standard GELU. This allows concluding that such quantization is not noticeable for stochastic optimization and can be applied without loss of quality and final properties of the model.

    In Fig.~\ref{fig:ru-dalle} we present training dynamic of ruDALL-E\footnote{Implementation is taken from https://github.com/sberbank-ai/ru-dalle} Malevich~\cite{Ramesh21} model on Russian Emoji dataset. The dataset~\cite{Shonenkov21} contains 2749 unique emoji icons and 1611 unique texts that were collected by web scrapping (the difference in quantities is due to the fact that there are sets, within which emojis differ only in color, moreover, some elements are homonyms in Russian). ruDALL-E Malevich is a big multimodal pretrained transformer, which learns the conditional distribution of images given some string of text (more precisely it autoregressively models the text and image tokens as a single stream of data). ruDALL-E Malevich encoder part is a 24 layer Transformer~\cite{Vaswani17} model with 16 attention heads, 2048 hidden dimensions and standard GELU nonlinearity, which in total has 1.3B parameters. It works with 128 text tokens, which are prepared from the text input using YTTM tokenizer\footnote{Implementation is taken from \url{https://github.com/VKCOM/YouTokenToMe}}, and 1024 image tokens, which are obtained after encoding the input image using Sber-VQGAN\footnote{Implementation is taken from \url{https://github.com/sberbank-ai/sber-vq-gan}}. Quantized backward for ruDALL-E Malevich shows same behaviour as for RoBERTa-base architecture: 1- and 2-bit versions, although coping with training perfectly fine, demonstrates minor performance degradation, while 3- and 4-bit versions are indistinguishable from the original GELU.

    \paragraph{ResNet Architecture.} To explore the influence of backward quantization on architectures other than Transformer and nonlinearities other than GELU, we trained ResNet18 model~\cite{He16} on ImageNet~\cite{Russakovsky15} benchmark~\cite{leclerc2022ffcv} dataset with ReLU replaced with Swish function ($\mathrm{Swish}(x) = x \sigma(x)$, where $\sigma$ is a sigmoid function) or with GELU, see Fig.~\ref{fig:resnet18-gelu-swish}. The behaviour of our quantization scheme remains the same for different network architectures and nonlinearities: 1- and 2- bits have minor performance drop, while 3- and 4- bits are on par with unchanged nonlinearity.

    \paragraph{ActNN.} As a baseline, we use another quantization scheme ActNN~\cite{Chen21}. It works in a much wider spectrum of situations, as it can quantize not only pointwise nonlinearity layers but also all kinds of linear layers (convolutional and dense layers), normalization layers and pooling layers. Without going deep into details, ActNN divides the saved tensor $H$ into chunks $\vb{h}_i$ where each chunk is of an equal size $G$. Then, given the quantization budget of $b$ bits, each chunk $\vb{h}_i$ is normalized: $\vb{u}_i = 2^b(h_i - \min\{\vb{h}_i\}) / (\max\{\vb{h}_i\} - \min\{\vb{h}_i\})$, and its randomly quantized version is saved $\vb{\bar{u}}_i = \lceil \vb{u}_i \rceil$ with prob. $\vb{u} - \lfloor \vb{u}_i \rfloor$, $\lfloor \vb{u}_i \rfloor$ otherwise. Random rounding is performed in order to guarantee that the quantization is unbiased. For each group, two additional values $\min\{\vb{h}_i\}$ and $\max\{\vb{h}_i\}$ are saved as well, but for the group size of $G = 256$ it is only $0.125$ additional bits per element, which we ignore in our following tests.
    ActNN by construction does not take into account the global behaviour of the nonlinearity derivative. We argue that for nonlinearity layers, it is very crucial, and thus our preoptimized quantization scheme is more preferable. To prove that, we consider ActNN behaviour on the QQP task from the GLUE benchmark with respect to different quantization budgets and compare it with our method. In Fig.~\ref{fig:actnn-our} train and validation losses are shown, and in Table~\ref{table:actnn-our} we report the final accuracy for the QQP task. In general, our method with 1 bit less budget works the same or better than ActNN.

    \begin{table}[H] \label{table:actnn-our}
    \centering
    \begin{tabular}{|l|l|l|l|l|}
        \hline
              & ActNN & Our \\ \hline
            1-bit   & $0.8880 \pm 0.0008$ & $0.9080 \pm 0.0006$ \\ \hline
            2-bit   & $0.9072 \pm 0.0005$ & $0.9097 \pm 0.0006$ \\ \hline
            3-bit   & $0.9106 \pm 0.0003$ & $0.9114 \pm 0.0007$ \\ \hline
            4-bit   & $0.9113 \pm 0.0006$ & $0.9112 \pm 0.0005$ \\ \hline

    \end{tabular}
    \end{table}

    \begin{figure*}[h]
        \centering
        \includegraphics[width=\textwidth]{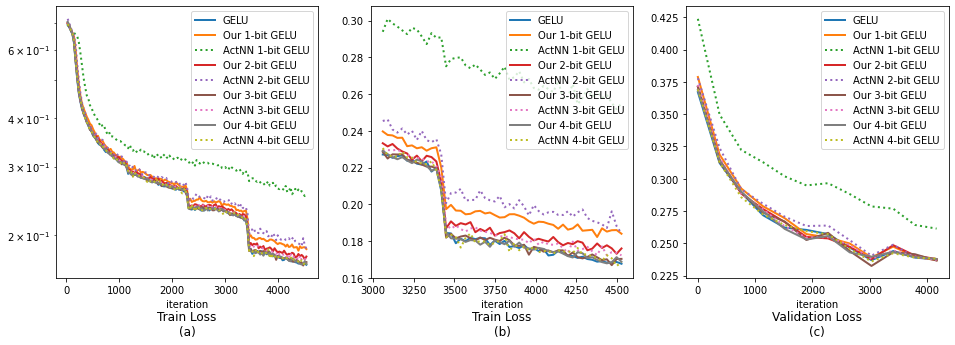}
        \caption{Comparison of RoBERTa-base on QQP task from GLUE benchmark with ActNN quantization and our quantization schemes. (a): Train loss. (b): Train loss during the last third of the training. (c): Validation loss. Averaged across ten runs.}
        \label{fig:actnn-our}
    \end{figure*}

    \paragraph{Memory saving.} Table~\ref{table:memory} shows memory measurements for different models with different number of bits per element for backward quantization. As was shown experimentally, for many tasks 3 bits is already enough for lossless training. In practice, fewer bits may be chosen to even further increase the batch size and consequently speed up training or to fit a larger model on a device that did not fit it before.

    \begin{table}[t] \label{table:glue-memory}
        \centering
        \caption{Peak memory usage in training time during fine-tuing on GLUE.}
        \label{tab:peak-memory-usage}
        \begin{tabular}{|lll|r|r|}
        \hline
        Task & Batch & BitWidth &  Mem, GiB &  Saving, \% \\
        \hline
        MRPC & 128 & Vanilla &     11.34 &        0.0 \\
             &     & 3-bit &      9.78 &       13.8 \\
             &     & 2-bit &      9.71 &       14.4 \\
        \hline
        QNLI & 16  & Vanilla &     11.73 &        0.0 \\
             &     & 3-bit &     10.77 &        8.2 \\
             &     & 2-bit &     10.75 &        8.4 \\
        \hline
        STS2 & 256 & Vanilla &     13.31 &        0.0 \\
             &     & 3-bit &     11.25 &       15.5 \\
             &     & 2-bit &     11.17 &       16.1 \\
        \hline
        \end{tabular}
    \end{table}

    \begin{table}[h] \label{table:memory}
    \centering
    \resizebox{.32\textwidth}{!}{%
    \begin{tabular}{|l|l|c|c|}
        \hline
        Model     & Quant & Saving & \makecell{Max Batch Size} \\ \hline

         \textbf{ResNet-101} & & & 131 \\
         256x256 size  & 1-bit & 30\% & 170 (+29.8\%) \\
            & 2-bit & 29\% & 169 (+29.0\%) \\
            & 3-bit & 28\% & 167 (+27.5\%) \\
            & 4-bit & 27\% & 165 (+26.0\%) \\
         \textbf{DenseNet-121} & & & 126 \\
         256x256 size  & 1-bit & 31\% & 165 (+31.0\%) \\
            & 2-bit & 30\% & 164 (+30.2\%) \\
            & 3-bit & 29\% & 162 (+28.6\%) \\
            & 4-bit & 28\% & 161 (+27.8\%) \\
         \textbf{Efficient Net B7} & & & 47 \\
         256x256 size  & 1-bit & 26\% & 59 (+25.5\%) \\
            & 2-bit & 25\% & 58 (+23.4\%) \\
            & 3-bit & 24\% & 58 (+23.4\%) \\
            & 4-bit & 23\% & 57 (+21.3\%) \\
         \textbf{RoBERTa-base} & & & 154 \\
         256 seq. len  & 1-bit & 16\% & 179 (+16.2\%) \\
            & 2-bit & 15\% & 178 (+15.6\%) \\
            & 3-bit & 15\% & 177 (+14.9\%) \\
            & 4-bit & 14\% & 176 (+14.3\%) \\
         \textbf{RoBERTa-large} & & & 54 \\
         256 seq. len  & 1-bit & 16\% & 63 (+16.7\%) \\
            & 2-bit & 16\% & 63 (+16.7\%) \\
            & 3-bit & 15\% & 62 (+14.8\%) \\
            & 4-bit & 15\% & 62 (+14.8\%) \\
         \textbf{GPT2} & & & 83 \\
         256 seq. len  & 1-bit & 42\% & 117 (+41.0\%) \\
            & 2-bit & 41\% & 116 (+39.8\%) \\
            & 3-bit & 39\% & 114 (+37.3\%) \\
            & 4-bit & 38\% & 113 (+36.1\%) \\
        \hline
    \end{tabular}
    }
    \caption{Memory savings and maximum batch size for popular models for different quantization budget. You can find more detailed version in Appendix~\ref{table:appendix-memory}}
    \end{table}

\section{Related Work}
    The reduction of the memory footprint is an important topic.
    To save memory during training, in addition to working with stored activations, the memory used to store model parameters can be compressed. Quantization~\cite{Bondarenko21,Bengio13,Banner19,Jacob18,Nagel21,Krishnamoorthi18} limits the admissible values of weights to some small finite set. Thus less memory is needed for storage. The low-rank representation of weights~\cite{Hrinchuk20,Phan20,Gusak19,gusak2021reduced,Cui20,Novikov18,Lebedev14} assumes some internal structure of model weights and saves memory by explicitly using this structure with low-rank methods from linear algebra. Low precision learning and low precision optimizers focus on using the lower precision floats to store weights, optimization parameters, and model gradients. All of these approaches are complementary to the proposed one and can be used together.

    Checkpointing~\cite{Beaumont19,Beaumont21,Chen16} methods save memory by the cost of more calculations. It stores a fewer number of activations and repeats the calculation of the rest from the saved checkpoints. Offloading methods~\cite{Beaumont20} send the saved activations to the computer's RAM and load them back to the video memory on the backwards passes, which also saves GPU memory at the cost of host-device communication time.

    ActNN~\cite{Chen21} is a framework for quantizing stored activations adaptively on the fly. In contrast to our work, it allows quantizing not only layers of element-by-element activations but also many others, including convolutional, normalization and linear layers. However, this method depends on the distribution of elements of quantizable tensors and, because of that, its performance may degrade. Our approach, on the other hand, selects data-agnostic optimal quantization, which in practice turns out to be sufficient and easier to use.

\section{Conclusion}
    We have proposed a method to reduce memory consumption during the training of deep neural network models by storing less information for backward pass in the element-wise activation functions. For effective training, there is no need to calculate the derivative of the activation functions precisely, but only its piecewise-constant approximation is sufficient. This makes it possible to save not the entire input tensor at each application of the activation function, but only the interval number in the piecewise-constant approximation. Experiments show that for a wide class of models and problems, storing only 3 bits of information per tensor element does not lead to degradation of the learning quality and saves about 20 percent of memory. We have proposed an efficient algorithm for constructing an optimal piecewise-constant approximation. The proposed drop-in replacements for popular activation functions (ReLU, GELU, Swish, Sigmoid and others) do not depend on the neural network model, the problem to be solved, or the peculiarities of data distribution. The replacement of the original activation functions by the proposed method can be performed at any training stage (both to models trained from scratch and to pre-trained models for subsequent fine-tuning) and does not require any changes in the training pipelines. An efficient CUDA implementation of the proposed method, together with pre-computed piecewise-constant approximations for many popular activation functions, is available for PyTorch at GitHub repository\footnote{Source code repository can be found at \url{https://github.com/SkoltechAI/fewbit}.}.

\section{Acknowledgements}

    The work was supported by the Analytical center under the RF Government (subsidy agreement 000000D730321P5Q0002, Grant No. 70-2021-00145 02.11.2021).

\bibliography{main}
\bibliographystyle{icml2022}

\newpage
\appendix
\onecolumn

\section{Detailed examples of few-bit approximations for popular nonlinearity layers}

\begin{figure}[H]
    \centering
    \includegraphics[height=\textheight]{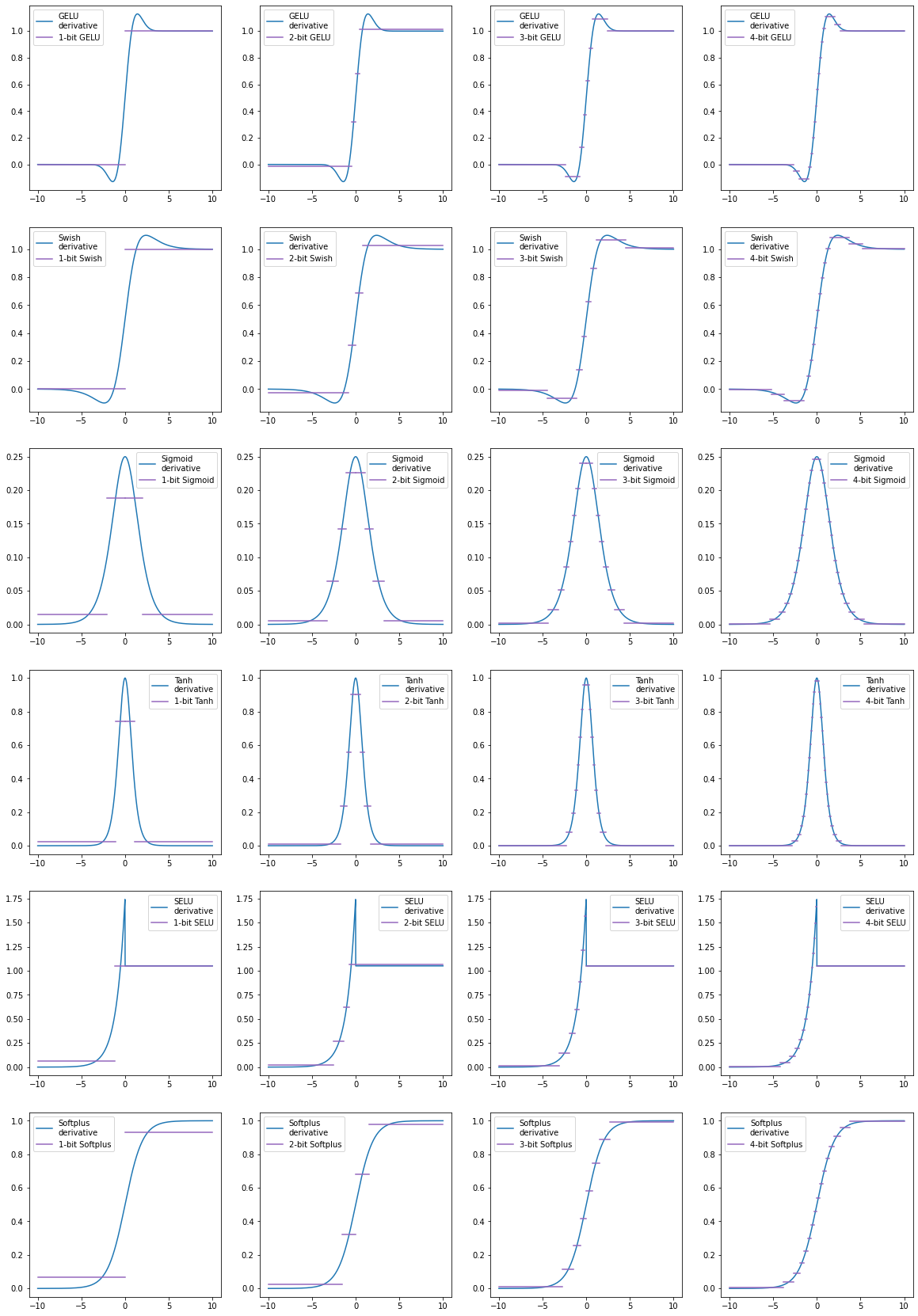}
    \caption{1- to 4-bit approximations of popular nonlinearty layers.}
    \label{fig:actnn-our}
\end{figure}

\section{Detailed memory measurements for different models}
    We provide memory measurements for different model architectures in Table~\ref{table:appendix-memory}. "Model size" is the total memory used for storing model parameters (without model gradients and optimizator statistics). "All activations size" is the total memory used by tensors, saved for backward pass. "Nonlinearity activations size" is the part of all activations used by nonlinearity layers. Maximum batch size is calculated with the assumption, that three model copies are stored on the device (model parameters, model gradients and optimizer statistics like weight moments in SGD with momentum).

    \begin{table}[H] \label{table:appendix-memory}
    \begin{tabular}{|l|c|c|c|c|c|c|c|}
        \hline
             & \makecell{Model \\ Size \\ (Mb)} & \makecell{All\\\small Activations\\Size\\(Mb)} & \makecell{\small Nonlinearity\\\small Activations\\Size\\(Mb)} & \makecell{1-bit\\Saving\\\small{Max batch size}} & \makecell{2-bit\\Saving\\\small{Max batch size}} & \makecell{3-bit\\Saving\\\small{Max batch size}} & \makecell{4-bit\\Saving\\\small{Max batch size}} \\ \hline

        \textbf{ResNet-18} & 44.6 & 40.0 & 11.5 & \makecell{28\% \\ 1010 (+27.5\%)} & \makecell{27\% \\ 1001 (+26.4\%)} & \makecell{26\% \\ 992 (+25.3\%)} & \makecell{25\% \\ 984 (+24.2\%)} \\ \hline
        \textbf{ResNet-50} & 99.2 & 156.8 & 47.9 & \makecell{30\% \\ 256 (+29.3\%)} & \makecell{29\% \\ 254 (+28.3\%)} & \makecell{28\% \\ 252 (+27.3\%)} & \makecell{27\% \\ 249 (+25.8\%)} \\ \hline
        \textbf{ResNet-101} & 171.4 & 234.5 & 73.4 & \makecell{30\% \\ 170 (+29.8\%)} & \makecell{29\% \\ 169 (+29.0\%)} & \makecell{28\% \\ 167 (+27.5\%)} & \makecell{27\% \\ 165 (+26.0\%)} \\ \hline
        \textbf{ResNet-152} & 232.3 & 328.2 & 104.9 & \makecell{31\% \\ 121 (+31.5\%)} & \makecell{30\% \\ 120 (+30.4\%)} & \makecell{29\% \\ 119 (+29.3\%)} & \makecell{28\% \\ 117 (+27.2\%)} \\ \hline
        \textbf{DenseNet-121} & 30.9 & 243.8 & 79.1 & \makecell{31\% \\ 165 (+31.0\%)} & \makecell{30\% \\ 164 (+30.2\%)} & \makecell{29\% \\ 162 (+28.6\%)} & \makecell{28\% \\ 161 (+27.8\%)} \\ \hline
        \textbf{DenseNet-161} & 112.4 & 457.2 & 145.3 & \makecell{31\% \\ 87 (+29.9\%)} & \makecell{30\% \\ 87 (+29.9\%)} & \makecell{29\% \\ 86 (+28.4\%)} & \makecell{28\% \\ 85 (+26.9\%)} \\ \hline
        \textbf{DenseNet-169} & 54.7 & 296.3 & 95.3 & \makecell{31\% \\ 136 (+30.8\%)} & \makecell{30\% \\ 134 (+28.8\%)} & \makecell{29\% \\ 133 (+27.9\%)} & \makecell{28\% \\ 132 (+26.9\%)} \\ \hline
        \textbf{DenseNet-201} & 77.4 & 382.2 & 123.9 & \makecell{31\% \\ 105 (+31.2\%)} & \makecell{30\% \\ 104 (+30.0\%)} & \makecell{29\% \\ 103 (+28.8\%)} & \makecell{28\% \\ 102 (+27.5\%)} \\ \hline
        \textbf{Efficient Net B0} & 20.4 & 112.4 & 32.4 & \makecell{28\% \\ 360 (+27.7\%)} & \makecell{27\% \\ 357 (+26.6\%)} & \makecell{26\% \\ 354 (+25.5\%)} & \makecell{25\% \\ 351 (+24.5\%)} \\ \hline
        \textbf{Efficient Net B3} & 47.5 & 218.6 & 59.5 & \makecell{26\% \\ 185 (+26.7\%)} & \makecell{26\% \\ 183 (+25.3\%)} & \makecell{25\% \\ 182 (+24.7\%)} & \makecell{24\% \\ 180 (+23.3\%)} \\ \hline
        \textbf{Efficient Net B7} & 256.3 & 673.5 & 178.9 & \makecell{26\% \\ 59 (+25.5\%)} & \makecell{25\% \\ 58 (+23.4\%)} & \makecell{24\% \\ 58 (+23.4\%)} & \makecell{23\% \\ 57 (+21.3\%)} \\ \hline
        \textbf{VGG 11} & 507.2 & 100.9 & 37.0 & \makecell{36\% \\ 386 (+35.4\%)} & \makecell{34\% \\ 382 (+34.0\%)} & \makecell{33\% \\ 377 (+32.3\%)} & \makecell{32\% \\ 373 (+30.9\%)} \\ \hline
        \textbf{VGG 16} & 528.2 & 163.8 & 68.5 & \makecell{41\% \\ 237 (+40.2\%)} & \makecell{39\% \\ 234 (+38.5\%)} & \makecell{38\% \\ 231 (+36.7\%)} & \makecell{37\% \\ 228 (+34.9\%)} \\ \hline
        \textbf{VGG 19} & 548.4 & 178.8 & 75.0 & \makecell{41\% \\ 217 (+40.9\%)} & \makecell{39\% \\ 214 (+39.0\%)} & \makecell{38\% \\ 211 (+37.0\%)} & \makecell{37\% \\ 208 (+35.1\%)} \\ \hline
        \textbf{RoBERTa-base} & 480.7 & 219.6 & 36.0 & \makecell{16\% \\ 179 (+16.2\%)} & \makecell{15\% \\ 178 (+15.6\%)} & \makecell{15\% \\ 177 (+14.9\%)} & \makecell{14\% \\ 176 (+14.3\%)} \\ \hline
        \textbf{RoBERTa-large} & 1355.6 & 578.1 & 96.0 & \makecell{16\% \\ 63 (+16.7\%)} & \makecell{16\% \\ 63 (+16.7\%)} & \makecell{15\% \\ 62 (+14.8\%)} & \makecell{15\% \\ 62 (+14.8\%)} \\ \hline
        \textbf{GPT2} & 491.0 & 331.1 & 144.0 & \makecell{42\% \\ 117 (+41.0\%)} & \makecell{41\% \\ 116 (+39.8\%)} & \makecell{39\% \\ 114 (+37.3\%)} & \makecell{38\% \\ 113 (+36.1\%)} \\ \hline
    \end{tabular}
    \end{table}

\end{document}